\documentclass[runningheads]{llncs}
\usepackage{graphicx}
\usepackage{dsfont}
\usepackage{esvect}
\usepackage[driverfallback=dvipdfm]{hyperref}
\usepackage{xcolor}

\begin{document}
\title{DAISI: Database for AI Surgical Instruction\thanks{This work was partly supported by the Office of the Assistant Secretary of Defense for Health Affairs under Award No. W81XWH-14-1-0042. Opinions, interpretations, conclusions and recommendations are those of the author and are not necessarily endorsed by the funders.}}
%
%

\author{Edgar Rojas-Mu\~noz\inst{1} \and
Kyle Couperus\inst{2} \and
Juan Wachs\inst{1}}

%
\authorrunning{E. Rojas-Mu\~noz et al.}
%

\institute{Purdue University, West Lafayette IN 47907, USA \and
Madigan Army Medical Center, Joint Base Lewis-McChord WA 98431, USA}
\maketitle              
\begin{abstract}
Telementoring surgeons as they perform surgery can be essential in the treatment of patients when in situ expertise is not available. Nonetheless, expert mentors are often unavailable to provide trainees with real-time medical guidance. When mentors are unavailable, a fallback autonomous mechanism should provide medical practitioners with the required guidance. However, AI/autonomous mentoring in medicine has been limited by the availability of generalizable prediction models, and  surgical procedures datasets to train those models with. This work presents the initial steps towards the development of an intelligent artificial system for autonomous medical mentoring. Specifically, we present the first Database for AI Surgical Instruction (DAISI). DAISI leverages on images and instructions to provide step-by-step demonstrations of how to perform procedures from various medical disciplines. The dataset was acquired from real surgical procedures and data from academic textbooks. We used DAISI to train an encoder-decoder neural network capable of predicting medical instructions given a current view of the surgery. Afterwards, the instructions predicted by the network were evaluated using cumulative BLEU scores and input from expert physicians. According to the BLEU scores, the predicted and ground truth instructions were as high as 67\% similar. Additionally, expert physicians subjectively assessed the algorithm using Likert scale, and considered that the predicted descriptions were related to the images. This work provides a baseline for AI algorithms to assist in autonomous medical mentoring.

\end{abstract}
\section{Introduction}
Telementoring surgeons as they perform surgery can be essential in rural, remote and even austere settings \cite{sebajang_role_2006,greenberg_surgical_2015}. Techniques such as telementoring through augmented reality and speech have been explored to provide general surgeons with remote supervision when no expert specialist is available on-site \cite{rojas-munoz_system_2020}. Nonetheless, all these approaches assume that there will always be a mentor readily available to provide medical guidance. However, when this is not the case \cite{bilgic_effectiveness_2017}, fallback autonomous mechanisms can provide medical practitioners with the necessary support. The creation of autonomous mentoring approaches based on artificial intelligence (AI) in medicine has been limited due to the lack of robust predicting models, and significant size and well curated datasets that can be used to train such models. Such datasets can include step-by-step demonstrations of how to perform surgical procedures from various medical specialties. 

\begin{figure}
\includegraphics[width=\textwidth]{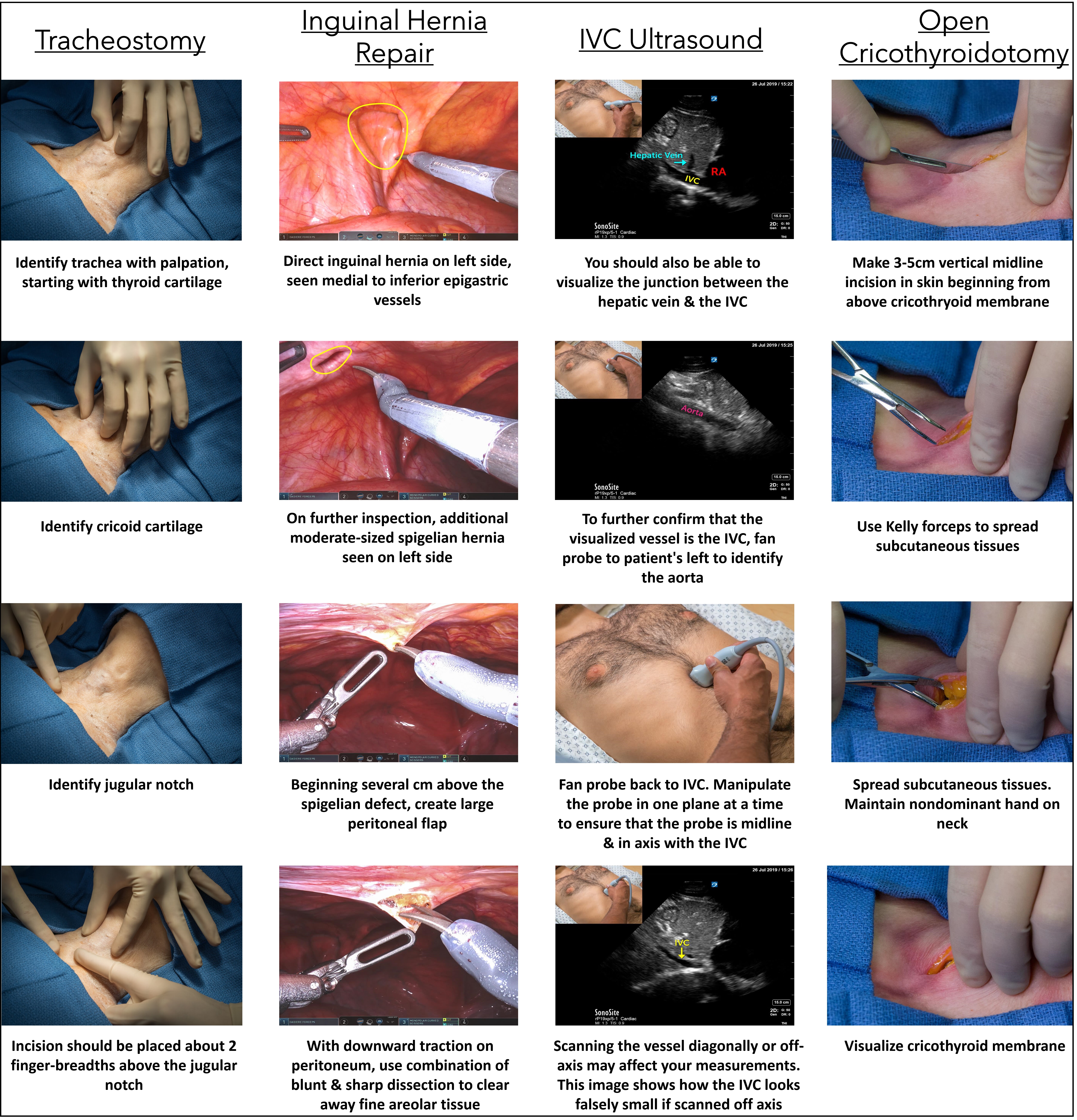}
\caption{Example of three different images and their associated textual descriptions from four different procedures in the DAISI database. The database includes images from 20 disciplines such as emergency medicine, and ultrasound-guided diagnosis.} \label{fig1}
\end{figure}

This works presents a Database for AI Surgical Instruction (DAISI; \href{https://engineering.purdue.edu/starproj/_daisi/}{\textcolor{blue}{Link to Database}}). DAISI provides step-by-step demonstrations of how to perform medical procedures. This is done by including images and text descriptions of procedures from 20 medical disciplines. Each image-text pair describes how to complete a step in the procedure. The database was created via input from 20 expert physicians from various medical centers, extracting data from academic medical textbooks related to the surgical technique, and acquiring imagery manually. Fig.~\ref{fig1} showcases images from four procedures in DAISI.

We evaluated the DAISI database by training a Neural Network (NN) to predict instructions. The model takes images of medical procedures as input, and outputs text descriptions of the instructions to perform. After training the NN using DAISI, the predicted instructions were evaluated using BLEU scores \cite{papineni_bleu_2002} and subjective input from expert physicians. The results presented in this work serve as a baseline for AI algorithms that can be used as surrogate human mentors. The paper proceeds as follows: Section 2 reviews approaches for medical autonomous guidance and related datasets. Section 3 describes DAISI and how it was used to train an AI model. Section 4 presents and discusses the results obtained from evaluating our AI model. Finally, Section 5 concludes the paper.

\section{Background}

Telementoring systems have been adopted to deliver expert assistance remotely \cite{kotwal_effect_2016}. Several studies demonstrate that such systems can provide surgeons with specialized assistance in austere settings when no expert is on-site \cite{sebajang_telementoring_2005,sebajang_role_2006}. However, telementoring platforms rely on remote specialists being available to assist, which often is not possible \cite{geng_addressing_2019}. A possible approach to convey guidance when no mentor is available is to incorporate AI into telementoring systems \cite{de_araujo_novaes_disruptive_2020}. When the mentor is not available, a virtual intelligent surrogate mentor can be activated.

AI algorithms have been previously explored as means to assist users during complex and time sensitive decision-making procedures \cite{cortes_agents_2007}. In the healthcare domain, for example, AI has been typically used in the diagnosis and prognosis of diseases \cite{patel_artificial_2020,mishra_automatic_2020}. Typically, the diagnosis is given via predictions of an AI model, trained using a database of medical records (e.g. radiology images, metabolic profiles). However, recent approaches have also incorporated AI into surgical instruction. The Virtual Operative Assistant is an example of an automated educational feedback platform \cite{mirchi_virtual_2020}. This platform was developed to provide automated feedback to neurosurgeons performing a virtual reality subpial brain tumor resection task. Other examples include AI to analyze surgical performance during virtual reality spine surgeries \cite{bissonnette_artificial_2019}, and integration with Augmented Reality to provide surgical navigation during surgery \cite{auloge_augmented_2019,jha_essence_2019}.

Recently, AI has been used to train models to predict medical image descriptions \cite{kisilev_medical_2016,singh_chest_2019,alsharid_captioning_2019,kougia_survey_2019}. For example, the ImageCLEFcaption challenge focuses on using AI to obtain text descriptions from radiology images \cite{pelka_overview_2019,lyndon_neural_2017,xu_concept_2019}. Similarly, IU X-RAY and PEIR GROSS are examples of public databases for radiology image captioning \cite{kougia_survey_2019,jing_automatic_2017}. These algorithms are used to describe the content images through captions. Conversely, such captions can be created to represent instructions in a task. Techniques for image captioning include template-based image captioning, retrieval-based image captioning, and novel caption generation \cite{hossain_comprehensive_2019}. Our work learns to generate visual-semantic correspondences between images and instructions using an architecture similar to \cite{karpathy_deep_2015}. In our case, however, the model is trained to generate instructions rather than descriptions.

\section{Methods}

The methodology to create such an AI surrogate mentor includes the creation of a curated dataset of medical images and their respective step-by-step descriptions. We demonstrate the use of such a dataset by training a Deep Learning (DL) framework which generates instructions from images. We evaluate our AI model and provide a benchmark for future AI surgical mentors.

\subsection{Creating a Database for AI Surgical Instruction}

The DAISI database contains 14586 color images and text descriptions of instructions to perform surgical procedures. DAISI contains one example for each of the 198 medical procedures from 20 medical disciplines including ultrasound-guided diagnosis, trauma and gynecology. The DAISI dataset is divided into training and testing sets. The training set contains 13232 images from 173 medical procedures, and the testing set contains 1354 images from 25 different medical procedures than those used in training. The database was made from: (a) medical images and instructions from the Thumbroll app \cite{thumbroll_llc_thumbroll_nodate}, a medical training app designed by physicians from Washington University School of Medicine, Stanford Health Care; UCLA, and University of Southern California. Overall, we acquired 14385 images with descriptions from various medical specialties (e.g. General Surgery, Internal Medicine), levels of medical training (e.g. clinical Medical Doctor trainee, senior Resident), and medical occupations (e.g. occupational therapy, osteopathic medicine); (b) Then, we extracted 125 images and captions from anatomy textbooks using \textit{PDFFigCapX} \cite{li_figure_2019}; (c) Lastly, we used a patient simulator (Tactical Casualty Care Simulator 1, Operative Experience) to acquire an addtional set of 76 images from procedures as chest needle decompression and intraosseous needle placement.

\subsection{Training an Intelligent Agent for Autonomous Mentoring}

\begin{figure}
\includegraphics[width=\textwidth]{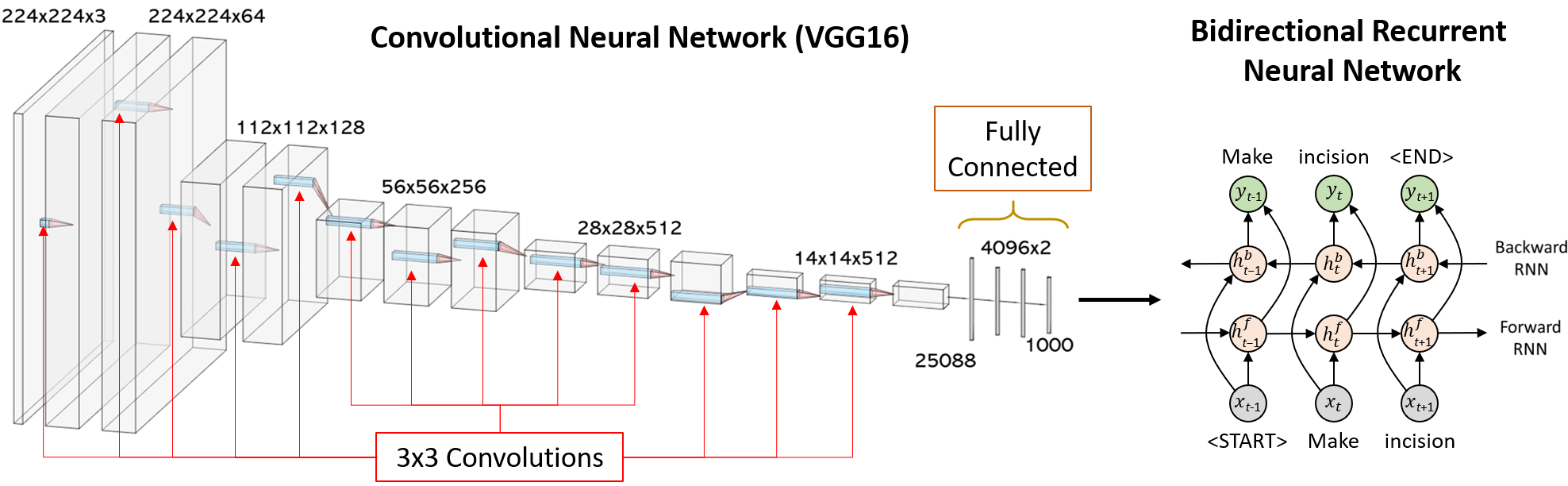}
\caption{Schematic of our encoder-decoder architecture. The CNN obtains vectors representing input images. These vectors are then used in the training of a BRNN that learns to predict surgical instructions. The VGG16 schematic was generated via \cite{lenail2019nn}.} \label{fig2}
\end{figure}

We used DAISI to train a DL model for autonomous mentoring. The algorithm receives images from medical procedures as input, and predicts an instruction associated with it. To generate text information from images, an encoder-decoder DL approach using a Convolutional Neural Network (CNN) and a Recursive Neural Network (RNN) was adopted. The CNN extracts and encodes visual features from the input images, and the RNN decodes these visual features into text descriptions (see Fig.~\ref{fig2}). 

Captioning techniques require a vocabulary containing the words appearing in the dataset at least $N$ times (defined by the \textit{Word Count} parameter). This constrains the words used to generate the instruction to a fix set. Our encoder-decoder architecture is based on NeuralTalk2 \cite{karpathy_deep_2015}. We use the VGG16 model as the encoder network \cite{simonyan_very_2014}. This model includes 13 convolutional layers with 5 pooling layers in-between. The convolutional layers use 3x3 convolutional filters to locate interest features in the images, and the pooling layers reduce the features' dimensionality. All hidden layers are equipped with Rectified Linear Units (ReLU). We performed cross validation using the Adam adaptive learning rate optimization to find individual learning rates for each parameter in the CNN \cite{kingma_adam_2014}. Finally, 4 fully connected layers are used to describe each image with a 1000-dimensional latent vector representation. We then use a Bidirectional Recurrent Neural Network (BRNN) as the decoder network to generate the text instructions \cite{schuster_bidirectional_1997}. The BRNN predicts instructions not only by receiving the CNN's final latent vector, but also by leveraging context around the word. This context is determined via forward and backward hidden states ($h_t^f$ and $h_t^b$, respectively) at each index $t$ ($t = 1$ \dots $T$), which denotes the position of a word in a sentence. The BRNN's formulation follows:
\begin{equation}
c_v = W_{hi} [{CNN}_\theta(Img)]
\end{equation}
\begin{equation}
h_t^f = ReLU(W_{hx}x_t + W_{hf} h_{t-1}^f + b_f + \mathds{1}(iter = 1) \bullet b_v)
\end{equation}
\begin{equation}
h_t^b = ReLU(W_{hx}x_t + W_{hb} h_{t+1}^b + b_b + \mathds{1}(iter = 1) \bullet b_v)
\end{equation}
\begin{equation}
y_t = Softmax(W_{ho}(h_t^f + h_t^b) + b_o)
\end{equation}

$W_{hi}$, $W_{hx}$, $W_{hf}$, $W_{hb}$, $W_{ho}$; $b_f$, $b_b$, and $b_o$ are the parameters and biases to be learnt by the model. ${CNN}_\theta(Img)$ CNN's final latent vector of the image $Img$. Thus, the image context vector $c_v$ provides the BRNN with information from the input image. This context vector $c_v$ is provided only during the first iteration ($iter = 1$), as suggested in \cite{karpathy_deep_2015}. The $x_t$ and $y_t$ vectors contain probabilities of each word in the vocabulary to be the word at the index $t$. The output vector $y_t$ is used as $x_{t+1}$ in the next iteration. In the first iteration, the output vector $y_t$ depends only on the context vector $c_v$, as $x_t$ takes a special initialization value (START) and $h_t^f$ and $h_t^b$ are initialized to 0. This formulation allows the model to predict more than one candidate instruction per image. The probability of each candidate being the correct instruction decreases for each additional prediction.

\subsection{Evaluating the Artificial Intelligent Mentor}

We evaluated our AI mentor using combinations of two parameters: \textit{Image Resolution}: High (1260x840), Medium (315x210), and Low (63x42), and \textit{Word Count}: 3, 5, and 7. Additionally, we conducted \textit{Inter-procedure} and \textit{Intra-procedure} evaluations. For the \textit{Inter-procedure} setting, the model had no prior information regarding the procedures in the test set. For the \textit{Intra-procedure} setting, a fraction of the images $P$ in the same procedure were assigned to the training set, while the rest remained in the test set. The test set consisted of every $\frac{1}{P}$ images from each procedure. In our case, $P$ was set to 0.5. While the \textit{Intra-procedure} setting reduced generalizability among procedures, it enhanced performance for procedures in the test set. To evaluate the algorithm's performance, the BLEU metric was computed between the predicted and the ground truth instructions. This is a state-of-the-art metric to evaluate text production models \cite{papineni_bleu_2002}. BLEU computes a 1-to-100 similarity score by comparing two sentences at the word \textit{n}-gram level. We report cumulative BLEU scores for 1-grams to 4-grams for the model’s top five candidate predictions, as they have reported correlations with human judgements \cite{ward_corpus-based_2002}. Finally, expert physicians evaluated the algorithm's performance subjectively. We randomly selected 16 images from the testing set and their predicted instructions. Afterwards, we used a survey to rate how related was each image to its predicted instruction. Each question in the survey included an image from a procedure, the name of the procedure; the instruction predicted, and a five scale ranking from: “Very Related” = 1, “Related” = 0.75, “Somewhat Related” = 0.5; “Not Related” = 0.25, and “Impossible to Tell” = 0.

\section{Results \& Discussion}

\begin{figure}
\includegraphics[width=\textwidth]{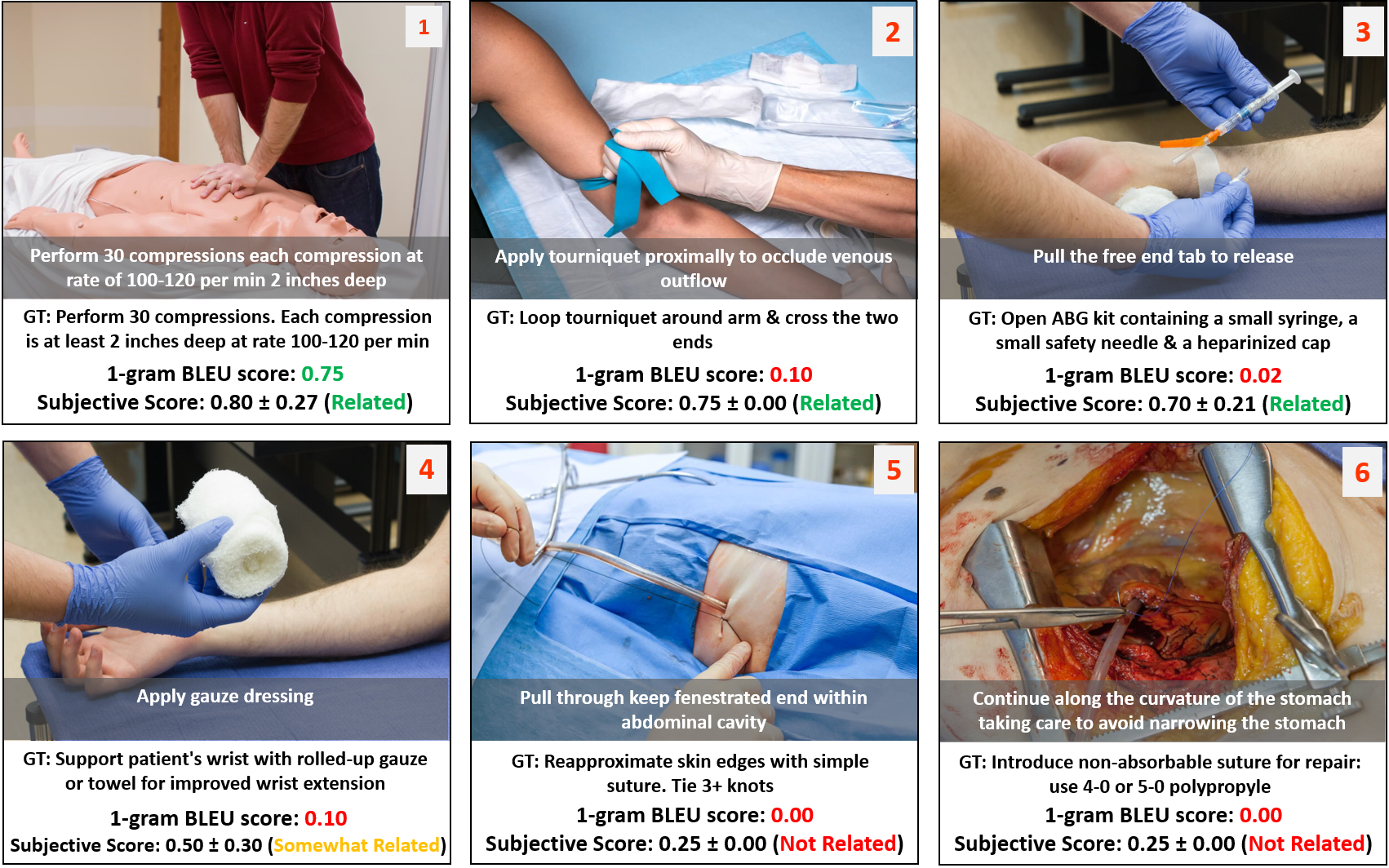}
\caption{Examples of instructions predicted by our AI model. The predicted instruction is in white font, inside the images. The ground truth (GT) instruction is written below. The 1-gram BLEU score and the score given by the physicians is included. High, average, and low scores are in green, yellow and red font, respectively.} \label{fig3}
\end{figure}

We validated our approach using four test folds. Fig.~\ref{fig3} presents instructions predicted by our AI model. The predicted instruction is written inside the images, whereas the ground truth instruction is written below. The data followed three main trends: (1) high BLEU and subjective scores (e.g. Fig.~\ref{fig3}, example 1); (2) low BLEU scores but high subjective scores (e.g. Fig.~\ref{fig3}, examples 2, 3 and 4); and (3) low BLEU and subjective scores (e.g. Fig.~\ref{fig3}, examples 5 and 6). The first trend are descriptions considered as correct predictions: they were similar to the ground truth and physicians considered them as adequate guidance. The second trend were descriptions that were not similar to the ground truth, but were consider as adequate guidance by the physicians. These descriptions included key elements from the image (e.g. tourniquet and gauze in Fig.~\ref{fig3}, examples 2 and 4, respectively), but did not use the phrasing of the ground truth. The third trend comprehends descriptions that were considered as incorrect predictions.

\begin{figure}
\includegraphics[width=\textwidth]{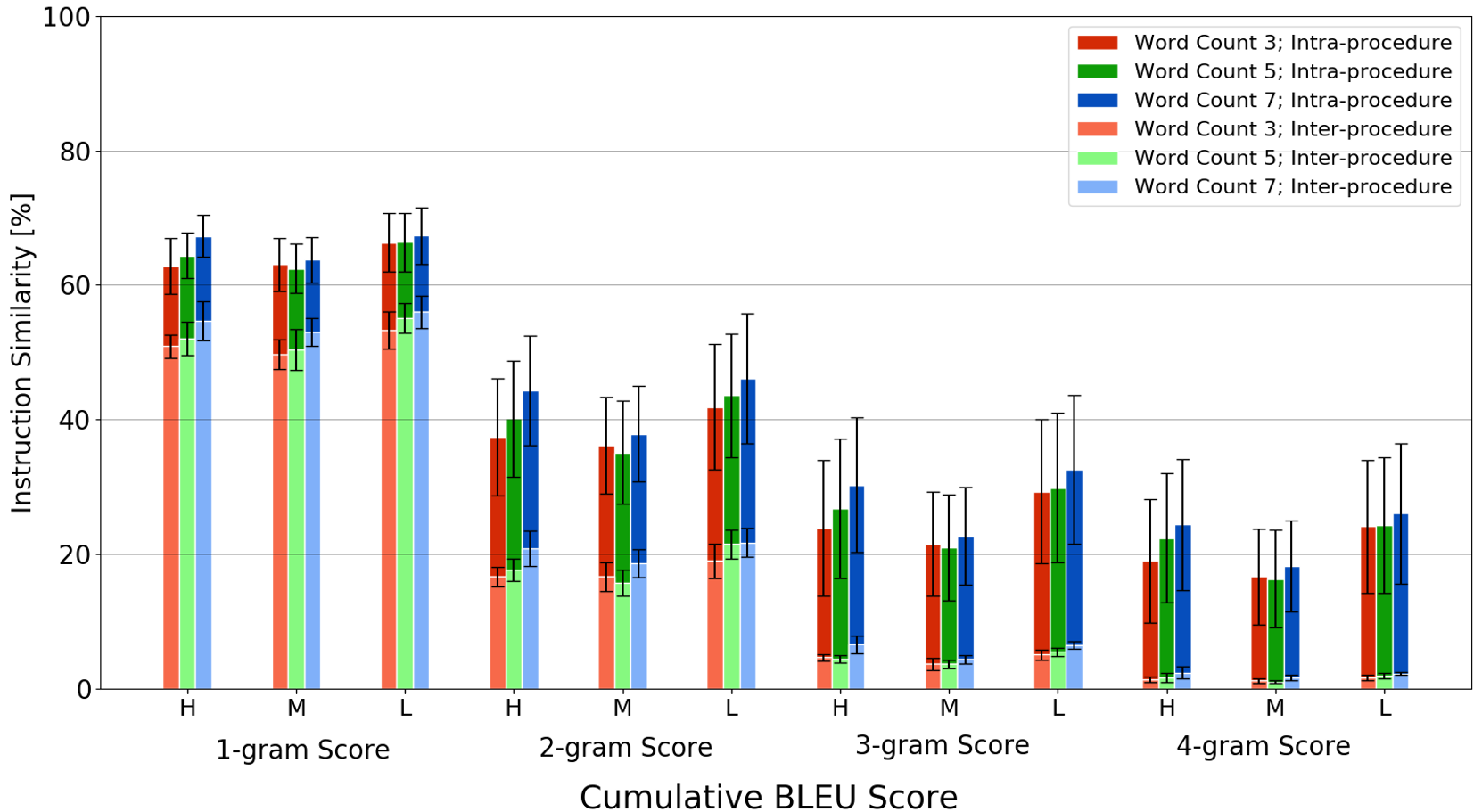}
\caption{Cumulative $n$-gram BLEU scores. Our model was evaluated using three \textit{Word Count} values (3, 5, 7), three \textit{Image Resolution} values (H=1260x840, M=315x210, L=63x42) and two testing approaches (\textit{Inter-procedure}, \textit{Intra-procedure}). The model obtained up to 67\% 1-gram BLEU scores, and up to 26\% 4-gram BLEU scores.} \label{fig4}
\end{figure}

Fig.~\ref{fig4} reports the cumulative BLEU scores for \textit{Inter-procedure} and \textit{Intra-procedure} testing. The captions predicted by our model obtained up to 67\% BLEU 1-gram and 26\% BLEU 4-gram scores. Our results surpassed those reported in state-of-the-art approaches for medical instructions prediction \cite{lyndon_neural_2017}. Overall, the BLEU scores were slightly higher for higher \textit{Word Count} values. A potential reason is that a reduced-size vocabulary increased the chance to learn meaningful relations between the images and the text descriptions. Likewise, the AI algorithm slightly favored smaller \textit{Image Resolution} values. While the best results reported are not high, our algorithm tackles a challenging problem due to the interclass variance among different medical procedures, which in turns has an impact the prediction capability of the network. As a reference value, the BLEU 1-gram score when comparing the ground truth instructions with descriptions constructed using random words from the vocabulary is less than 0.1\%. Our results show an improvement of over 4 folds over random guess.

Five expert physicians completed our subjective evaluation, for a total 80 responses. The physicians reported having $11.2 \pm 3.3$ years of medical expertise. On average, the physicians considered the predicted instructions to be “Somewhat Related” to the medical images ($0.51 \pm 0.32$). While this is an encouraging result, a drastic improvement is still required for useful AI mentoring for surgery. Therefore, the main value of this work is offering a baseline for future autonomous medical mentoring applications. The DAISI open dataset is a useful tool that allows the AI community to train machine learning models that learn clinical instructions. Future work includes adding more repetitions per procedure. While our \textit{Intra-procedure} testing approach alleviates this limitation, more repetitions can improve the prediction results. Finally, data augmentation techniques can be used to increase the size of the dataset, for example by training a Generative Adversarial Neural network \cite{frid-adar_gan-based_2018} to create new images and descriptions.

\section{Conclusion}

This work presented DAISI, a dataset to train AI algorithms that can act as surrogate surgical mentors by generating detailed surgical instructions. To evaluate DAISI as a knowledge base, an encoder-decoder neural network was train to predict surgical instructions. The instructions predicted by the network were evaluated using cumulative BLEU scores and input from expert physicians. According to the BLEU scores, the predicted and ground truth instructions were as high as 67\% similar. Moreover, expert physicians considered that randomly selected images and their predicted descriptions were related. The results from this work serve as a baseline for future AI algorithms assisting in autonomous medical mentoring.

\bibliographystyle{splncs04}
\bibliography{MICCAI_final.bib}

\end{document}